\documentclass[10pt,wcp]{jmlr}

\usepackage[T1]{fontenc}
\usepackage[utf8]{inputenc}

\usepackage{longtable}

\usepackage{booktabs}

\usepackage{mathtools}

\usepackage{algorithm2e}
\usepackage{algorithmic}

\usepackage{makecell}
\usepackage{paralist}
\usepackage{enumitem}
\usepackage{colortbl}
\usepackage{capt-of}
\usepackage{booktabs}
\usepackage{setspace}
\usepackage{multirow}
\SetKwComment{Comment}{$\triangleright$\ }{}

\jmlryear{2019}

\title[Parameter Estimation with the Ordered $\ell_{2}$ Regularization via an ADMM]{Parameter Estimation with the Ordered $\ell_{2}$ Regularization via an Alternating Direction Method of Multipliers
}

  \author{\Name{Mahammad Humayoo} \Email{humayoo@ict.ac.cn}\and
   \Name{Xueqi Cheng} \Email{cxq@ict.ac.cn}\\
   \addr CAS Key Laboratory of Network Data Science \& Technology, Institute of Computing Technology, Chinese Academy of Sciences, Beijing 100190, China\\
   \addr University of Chinese Academy of Sciences, Beijing 100049, China
   }

\begin{document}

\maketitle

\begin{abstract}
Regularization is a~popular technique in machine learning for model estimation and for avoiding overfitting.~Prior studies have found that modern ordered regularization can be more effective in handling highly correlated, high-dimensional data than traditional regularization. The~reason stems from the fact that the ordered regularization can reject irrelevant variables and yield an~accurate estimation of the parameters. How to scale up the ordered regularization problems when facing large-scale training data remains an~unanswered question. This paper explores the problem of parameter estimation with the ordered $\ell_{2}$-regularization via Alternating Direction Method of Multipliers (ADMM), called ADMM-O$\ell_{2}$. The advantages of ADMM-O$\ell_{2}$ include (i)~scaling up the ordered $\ell_{2}$ to a~large-scale dataset, (ii) predicting parameters correctly by excluding irrelevant variables automatically, and~(iii) having a~fast convergence rate. Experimental results on both synthetic data and real data indicate that ADMM-O$\ell_{2}$ can perform better than or comparable to several state-of-the-art baselines.
\end{abstract}

\section{Introduction}
\label{introduction}
In the machine learning literature, one of the most important challenges involves estimating parameters accurately and selecting relevant variables from highly correlated, high-dimensional data. Researchers have noticed many highly correlated features in high-dimensional data~\citep{tibshirani1996regression}. Models often overfit or underfit high-dimensional data because they have a~large number of variables but only a~few of them are actually relevant; most others are irrelevant or redundant. An underfitting model contributes to estimation bias (i.e., high bias and low variance) in the model fitting because it keeps out relevant variables, whereas an~overfitting raises estimation error (i.e., low bias and high variance) since it includes irrelevant variables in the model.\par
To illustrate an~application of our proposed method, consider a~study of gene expression data. This is a~high-dimensional dataset and contains highly correlated genes. The geneticist always likes to determine which variants/genes contribute to changes in biological phenomena (e.g., increases in blood cholesterol level, etc.)~\citep{bogdan2015slope}. Therefore, the~aim is to explicitly identify all relevant variants. The~penalized regularization models such as $\ell_{1}$, $\ell_{2}$, and~so forth have recently become a~topic of great interest within machine learning, statistics~\citep{tibshirani1996regression}, and~optimization~\citep{bach2012optimization} communities as classic approaches to estimate parameters.~The $\ell_{1}$-based method is not a~preferred selection method for groups of variables among which pairwise correlations are significant because the lasso arbitrarily selects a~single variable from the group without any consideration of which one to select~\citep{efron2004least}. Furthermore, if the selected value of a~parameter is too small, the~$\ell_{1}$-based method would select many irrelevant variables, thus degrading its performance. On the other hand, a~large value of the parameter would yield a~large bias~\citep{bogdan2013statistical}. Another point worth noting is that few $\ell_{1}$ regularization methods are either adaptive, computationally tractable, or distributed, but no such method contains all three properties together. Therefore, the~aim of this study is to develop a~model for parameter estimation and to determine relevant variables in highly correlated, high-dimensional data based on the ordered $\ell_{2}$. This model has the following three properties all together: adaptive (Our method is adaptive in the sense that it reduces the cost of including new relevant variables as more variables are added to the model due to rank-based penalization properties.), tractable (A computationally intractable method is a~computer algorithm that takes a~very long time to execute a~mathematical solution. A computationally tractable method is exactly the opposite of intractable method.), and~distributed.\par

Several adaptive and nonadaptive methods have been proposed for parameter estimation and variable selection in large-scale datasets. Different principles are adopted in these procedures to estimate parameters. For example, an~adaptive solution (i.e., an~ordered $\ell_{1}$)~\citep{bogdan2013statistical} is a~norm and, therefore, convex. Regularization parameters are sorted in non-increasing order in the ordered $\ell_{1}$, in which the ordered regularization penalizes regression coefficients according to their order, with higher orders closer to the top and having larger penalties. \cite{pan2017robust} proposed a~partial sorted $\ell_{p}$ norm, which is non-convex and non-smooth. In contrast, the~ordered $\ell_{2}$ regularization is convex and smooth, just~as the standard $\ell_{2}$ norm is convex and smooth in Reference~\citep{Azghani2015FastMM}. \cite{pan2017robust} considered p-values between $0<p\leq1$ that do not cover $\ell_{2}$, $\ell_{\infty}$ norms, and~so forth. \cite{pan2017robust} also did not provide details of other partially sorted norms when $p\geq2$ and used random projection and the partial sorted $\ell_{p}$ norm to complete the parameter estimation, whereas we have used ADMM with the ordered $\ell_{2}$. A~nonadaptive solution (i.e., an~elastic net)~\citep{zou2005regularization} is a~mixture of both ordinary $\ell_{1}$ and $\ell_{2}$. In particular, it~is a~useful model when the number of predictors (p) is much larger than the number of observations (n) or in any situation where the predictor variables are correlated.\par

Table \ref{tab:mot} presents the important properties of the regularizers. As seen in Table \ref{tab:mot}, $\ell_{2}$ and the ordered $\ell_{2}$ regularizers are suitable methods for highly correlated, high-dimensional grouping data rather than $\ell_{1}$ and the ordered $\ell_{1}$ regularizers. The ordered $\ell_{2}$ encourages grouping, whereas most of $\ell_{1}$-based methods promote sparsity.~Here,~grouping signifies a~group of strongly correlated variables in high-dimensional data. We used the ordered $\ell_{2}$ regularization in our method instead of $\ell_{2}$ regularization because the ordered $\ell_{2}$ regularization is adaptive. Finally, ADMM has a~parallel behavior for solving large-scale convex optimization problems. Our model also employs ADMM and inherits distributed properties of native ADMM~\citep{boyd2011distributed}. Hence, our model is also distributed. \cite{bogdan2013statistical} did not provide any details about how they applied ADMM in the ordered $\ell_{1}$ regularization.

In this paper, we propose ``Parameter Estimation with the Ordered $\ell_{2}$ Regularization via ADMM'' called ADMM-O$\ell_{2}$ to find relevant parameters from a~model. $\ell_{2}$ is a~ridge regression; similarly, the~ordered $\ell_{2}$ becomes an~ordered ridge regression. The main contribution of this paper is not to present a~superior method but rather to introduce a~quasi-version of the $\ell_{2}$ regularization method and to concurrently raise awareness of the existing methods. As part of this research, we introduced a~modern ordered $\ell_{2}$ regularization method and proved that the square root of the ordered $\ell_{2}$ is a~norm and, thus, convex. Therefore, it is also tractable. In addition, the~regularization method used an~ordered elastic net method to combine the widely used ordered $\ell_{1}$ penalty with modern ordered $\ell_{2}$ penalty for ridge regression. The ordered elastic net is also proposed by the scholars in this paper. To the best of our knowledge, this is one of the first method to use the ordered $\ell_{2}$ regularization with ADMM for parameter estimation and variable selection. In Sections \ref{or} and \ref{ADMMorr}, we explain the integration of ADMM with the ordered $\ell_{2}$ and further details about it.\par

The rest of the paper is arranged as follows. Related works are discussed in Section \ref{relatedwork}, along~with a~presentation of the ordered $\ell_{2}$ regularization in Section \ref{or}. Section \ref{ADMMorr} describes the application of ADMM to the ordered $\ell_{2}$. Section \ref{experiment} presents the experiments conducted. Finally, Section \ref{conclusions} closes the paper with a~conclusion.

\begin{table}
  \centering
  \caption{Properties of the different regularizers: Correlation is a~method that shows how strong the relationship between explanatory variables is. In the correlation column, ``Yes'' means a~strong correlation between variables and ``No'' means weaker (or no) correlation between variables. Stable means that estimates of the $\ell_{2}$-based methods are more stable when the explanatory variables are strongly correlated.}
  \label{tab:mot}
\scalebox{.7}[.7]{\begin{tabular}{lllllllll}
\toprule
\textbf{Regularizers} & \textbf{Promoting} & \textbf{Convex} & \textbf{Smooth} & \textbf{Adaptive} & \textbf{Tractable} & \textbf{Correlation} & \textbf{Stable} \\
\midrule
\makecell{$\ell_{0}$~\cite{daducci2014sparse,gong2013general}} & Sparsity & No & No & No & No & No & No               \\
\makecell{$\ell_{1}$~\cite{james2013introduction}} & Sparsity & Yes & No & No & Yes & No & No \\
\makecell{The Ordered $\ell_{1}$~\cite{bogdan2013statistical}} & Sparsity & Yes & No & Yes & Yes & No & No \\
\makecell{$\ell_{2}$~\cite{Wang2008HybridHS}} & Grouping & Yes & Yes & No & Yes & Yes & Yes \\
\makecell{The Ordered $\ell_{2}$\\}& Grouping & Yes & Yes & Yes & Yes & Yes & Yes \\
\makecell{Partial Sorted $\ell_{p}$~\cite{pan2017robust}}& Sparsity & No & No & Yes & No & No & No \\
 \bottomrule
\end{tabular}}
\end{table}

\section{Related Work}\label{relatedwork}
\vspace{-6pt}
\subsection{$\ell_{1}$ and $\ell_{2}$ Regularization}
\cite{deng2013group} presented efficient algorithms for group sparse optimization with mixed $\ell_{2,1}$ regularization for the estimation and reconstruction of signals. Their technique is rooted in a~variable splitting strategy and ADMM.~\cite{zou2005regularization} suggested that an~elastic net, a~generalization of the lasso, is a~linear combination of both $\ell_{1}$ and $\ell_{2}$ norm. It contributes to sparsity without permitting a~coefficient to become too large. However, \cite{candes2007dantzig} introduced a~new estimator called the Dantzig selector for a~linear model when the parameters are larger than the number of observations for which they established optimal $\ell_{2}$ rate properties under a~sparsity. \cite{chen2015fast} enforced sparse embedding to ridge regression, obtaining solutions $\tilde{x}$ with $\|\tilde{x} - x^{*}\|_{2}\leq\epsilon\|x^{*}\|$ small, where $x^{*}$ is optimal, and~also did this in $\mathcal{O}(nnz(A)+ n^{3}/\epsilon^{2})$ time, where nnz(A) is the number of nonzero entries of A. Recently, \cite{bogdan2013statistical} have proposed an~ordered $\ell_{1}$-regularization technique inspired from a~statistical viewpoint, in particular, by a~focus on controlling the false discovery rate (FDR) for variable selection in linear regressions. Our proposed method is similar but focused on parameter estimation based on the ordered $\ell_{2}$ regularization and ADMM. Several methods have been proposed based on Reference~\citep{bogdan2013statistical} and similar ideas. For example, \cite{bogdan2015slope} introduced a~new model-fitting strategy called Sorted L-One Penalized Estimation (SLOPE) which regularizes least-squares estimates with rank-dependent penalty coefficients. \cite{zeng2014decreasing} proposed DWSL1 as a~generalization of octagonal shrinkage and clustering algorithm (OSCAR) that aims to promote feature grouping without previous knowledge of the group structure. \cite{pan2017robust} have introduced an~image restoration method based on a~random projection and a~partial sorted $\ell_{p}$ norm. In this method, an~input signal is decomposed into two components: a~low-rank component and a~sparse component. The low-rank component is approximated by random projection, and~the sparse one is recovered by the partial sorted $\ell_{p}$ norm. Our method can potentially be used in various other domains such as cyber security~\citep{albanese2014recognizing} and recommendation~\citep{amato2017recommendation}.
\subsection{ADMM}
Researchers have paid a~significant amount of attention to ADMM because of its capability of dealing with objective functions independently and simultaneously. Second, ADMM has proved to be a~genuine fit in the field of large-scale data-distributed optimization. However, ADMM is not a~new algorithm; it was first introduced by References~\citep{glowinski1975approximation,gabay1976dual} in the mid-1970s, with roots as far back as the mid-1950s. In~addition, ADMM originated from an~augmented method with a~Lagrangian multiplier~\citep{hestenes1969multiplier}. It~became more popular when \cite{boyd2011distributed} published papers about ADMM. The~classic ADMM algorithm applies to the following “ADMM-ready” form of problems.
\vspace{-6pt}
\begin{align}
\begin{cases}
\label{eq1:1}
minimize    \quad f(x) + g(z) \\ s.t. \quad   Ax + Bz = c
\end{cases}
\end{align}

The wide range of applications have also inspired the study of the convergence properties of ADMM. Under mild assumptions, ADMM can converge for all choices of the step size.~\cite{ghadimi2015optimal} provided some advice on tuning over-relaxed ADMM in the quadratic problems. \cite{deng2016global} have also suggested linear convergence results under the consideration of only a~single strongly convex term, given that linear operator’s A and B are full-rank matrices. These~convergence results bound error as measured by an~approximation to the primal–dual gap. \cite{goldstein2014fast} created an~accelerated version of ADMM that converges more quickly than traditional ADMM under an~assumption that both objective functions are strongly convex. \cite{yan2016self} explained, in detail, the~different kinds of convergence properties of ADMM and their prerequisite rules for converging. For further studies on ADMM, see Reference~\citep{boyd2011distributed}.

\section{Ridge Regression with the Ordered $\ell_{2}$ Regularization}\label{or}
\vspace{-6pt}
\subsection{The Ordered $\ell_{2}$ Regularization}\label{ol2}
The proposed parameter estimation and variable selection method in this paper is computationally manageable and adaptive.~This procedure depends on the ordered $\ell_{2}$ regularization. Let $\lambda = (\lambda_{1},\lambda_{2},\dots,\lambda_{p})$ be a~decreasing sequence of positive scalars that satisfy the following condition:
\begin{align}
\label{lambdaseq}
\lambda_{1}\geq\lambda_{2}\geq\lambda_{3}\geq ... \geq\lambda_{p}\geq0
\end{align}

The ordered $\ell_{2}$ regularization of a~vector $x\in\mathbb{R}^{p}$ when $\lambda_{1}>0$ can be defined as follows:
\begin{align}
\label{lfunc}
J_{\lambda}(x) = \lambda_{1} x_{(1)}^{2} + \lambda_{2} x_{(2)}^{2} + \dots+ \lambda_{p} x_{(p)}^{2} = \substack{p \\ \sum \\ k=1}\lambda_{BH^{(k)}} x_{(k)}^{2}
\end{align}
where $\lambda_{BH^{(k)}}$ is called a~BHq method~\citep{benjamini1995controlling}, which generates an~adaptive and a~non-increasing value for $\lambda$ (Reference~\cite{bogdan2015slope}, Section 1.1). The details of $\lambda_{BH^{(k)}}$ are available in Section \ref{Over-relaxedform}. For ease of presentation, we~have written $\lambda_{k}$ in place of $\lambda_{BH^{(k)}}$ in the rest of the paper. $x_{(1)}^{2}\geq x_{(2)}^{2}\geq x_{(3)}^{2}\geq \dots \geq x_{(p)}^{2}$  is the order statistic of the magnitudes of $x$~\citep{davidorder}. The subscript $k$ of $x$ enclosed in parentheses indicates the $k$th-order statistic of a~sample. Suppose that $x$ is a~sample size of 4.  Hence, four numbers are observed in $x$ if the sample values of $x$ = ($-$2.1, $-$0.5, 3.2, 7.2). The order statistics of $x$ would be $x_{(1)}^{2} = 7.2^{2}, x_{(2)}^{2} = 3.2^{2}, x_{(3)}^{2} = 2.1^{2}, x_{(4)}^{2} = 0.5^{2}$. The ordered $\ell_{2}$ regularization is expressed as the first largest value of $\lambda$ times the square of the first largest entry of $x$, plus the second largest value of $\lambda$ times the square of the second largest entry of $x$, and~so on. $A \in \mathbb{R}^{n \times p}$ and $b \in \mathbb{R}^{n}$ are a~matrix and a~vector, respectively. The ordered $\ell_{2}$ regularized loss minimization can be expressed as follows:
\begin{equation}\label{I1}
\begin{split}
\substack{min \\ x \in\ \mathbb{R}^{p}} \frac{1}{2}\|Ax - b\|_{2}^{2} + \frac{1}{2}\{ \lambda_{1} x_{(1)}^{2} + \dots+ \lambda_{p} x_{(p)}^{2}\}
\end{split}
\end{equation}

\begin{theorem}
\label{theorem1}
If the square root of $J_{\lambda}(x)$ (Equation (\ref{lfunc})) is a~norm on $\mathbb{R}^{p}$ and a~function $\|.\|: \mathbb{R}^{p}\rightarrow \mathbb{R}$ satisfying the following three properties, then the following Corollaries 1 and 2 are true. 
\begin{enumerate}[leftmargin=*,labelsep=4.9mm,label=\roman*]
\item (Positivity) $\|x\|\geq0$ for any $x\in\mathbb{R}^{p}$ and $\|x\|=0$ if and only if $x=0$.
\item (Homogeneity) $\|cx\|=|c|\|x\|$ for any $x\in\mathbb{R}^{p}$ and $c\in\mathbb{R}$.
\item (Triangle inequality) $\|x+y\|\leq\|x\|+\|y\|$ for any $x,y\in\mathbb{R}^{p}$.
\end{enumerate}

Note: $\|x\|$ and $\|x\|_{2}\ are\ used\ interchangeably$.
\end{theorem}
\begin{corollary}
\label{coroll1}
When all the $\lambda_{k}'s$ take on an~equal positive value, $J_{\lambda}(x)$ reduces to the square of the usual $\ell_{2}$ norm.
\end{corollary}
\begin{corollary}
\label{coroll2}
When $\lambda_{1}>0$ and $\lambda_{2}$ = \dots = $\lambda_{p}=0$, the~square root of $J_{\lambda}(x)$ reduces to the $\ell_{\infty}$ norm.
\end{corollary}

Proofs of the theorem and corollaries are provided in Appendix \ref{proofs}. Table \ref{tab:table1} shows the notations used in this paper and their meanings.
\begin{table}
  \centering
  \caption{Notations and explanations.}
  \label{tab:table1}
  \scalebox{0.95}[0.95]{\begin{tabular}{llll}
\toprule
\textbf{Notations}   & \textbf{Explanations} & \textbf{Notations} & \textbf{Explanations} \\
\midrule
Matrix      & denoted by uppercase letter  & f & loss convex function \\
Vector      & denoted by lowercase letter  & g & Regularizer part ($\ell_{1}$ or $\ell_{2}$ etc.) \\
$\|.\|_{1}$ & $\ell_{1}$ norm & $\partial f(x)$ & subdifferential of convex function f at x \\
$\|.\|_{2}$       & $\ell_{2}$ norm & $\partial g(z)$ & subdifferential of convex function g at z \\
$\|x\|_{1} = \substack{n \\ \sum \\ k=1}|x_{k}|$ & $\ell_{1}$ norm   & L1 & the $\ell_{1}$ norm or the lasso \\
$\|x\|_{2} = \sqrt{\substack{n \\ \sum \\ k=1}|x_{k}|^{2}}$ & $\ell_{2}$ norm  & OL1 & ordered $\ell_{1}$ norm or ordered lasso \\
$\|x\|^{2}_{2} = \substack{n \\ \sum \\ k=1}|x_{k}|^{2}$ & square of $\ell_{2}$ norm & OL2 & ordered $\ell_{2}$ norm or ordered ridge regression \\
$J_{\lambda}(.)$ & the ordered norm & Eq. & equation \\
\bottomrule
\end{tabular}}
\begin{tabular}{llll}
\multicolumn{1}{c}{\footnotesize \textbf{Note:} We often use the ordered $\ell_{2}$ norm/regularization, OL2, and~ADMM-O$\ell_{2}$ interchangeably.}   
\end{tabular}
\end{table}

\subsection{The Ordered Ridge Regression}
\label{orr}
We propose an~ordered ridge regression in Equation (\ref{p2}) and call it the ordered ridge regression because we used an~ordered $\ell_{2}$ regularization with an~objective function instead of a~standard $\ell_{2}$ regularization. The ordered ridge regression is commonly used for parameter estimation and variable selection, particularly when data are strongly correlated and highly dimensional. The ordered ridge regression can be defined as follows:
\begin{equation}\label{p2}
\begin{split}
\! \substack{min \\ x\ \in\ \mathbb{R}^{p}}\frac{1}{2}\|Ax - b\|_{2}^{2} + \frac{1}{2} J_{\lambda}(x)\\
 = \substack{min \\ x\ \in\ \mathbb{R}^{p}}\frac{1}{2}\|Ax - b\|_{2}^{2} + \frac{1}{2} \substack{p \\ \sum \\ k=1}\lambda_{k}|x_{(k)}|^{2}
\end{split}
\end{equation}
where $x\in \mathbb{R}^{p}$ denotes an~unknown regression coefficient, $A \in \mathbb{R}^{n \times p}(p \gg n)$ is a~known matrix, $\emph{b}\in \mathbb{R}^{n}$ represents a~response vector, and~$J_{\lambda}(x)$ is the ordered $\ell_{2}$ regularization. The optimal parameter choice for the ordered ridge regression is much more stable than that for a~regular lasso; also, it achieves adaptivity in the following senses.
\begin{enumerate}[leftmargin=*,labelsep=4.2mm, label=\roman*]
\item For decreasing $(\lambda_{k})$, each parameter $\lambda_{k}$ marks the entry or removal of some variable from the current model (therefore, its coefficient becomes either zero or nonzero); thus, coefficients remain constant in the model. We achieved this by putting some threshold values for $\lambda_{k}$ (Reference~\cite{bogdan2013statistical}, Section~1.4).
\item We observed that the price of including new variables declines as more variables are added to the model when the $\lambda_{k}$ decreases.
\end{enumerate}

\section{Applying ADMM to the Ordered Ridge Regression}
\label{ADMMorr}
In order to apply ADMM to the problem in Equation (\ref{p2}), we first transform it into an~equivalent form of the problem in Equation (\ref{eq1:1}) by introducing an~auxiliary variable $z$.
\begin{equation}\label{p3}
\begin{split}
\substack{min \\ x,z\ \in\ \mathbb{R}^{p}}  \frac{1}{2}\|Ax - b\|_{2}^{2} + \frac{1}{2} J_{\lambda}(z) \quad s.t. \quad x-z = 0
\end{split}
\end{equation}

We can see that Equation (\ref{p3}) has two blocks of variables (i.e., $x$ and $z$). Its objective function is separable in the form of Equation (\ref{eq1:1}) since $f(x) = \frac{1}{2}\|Ax - b\|_{2}^{2}$ and $g(z) = \frac{1}{2} J_{\lambda}(z) = \frac{1}{2} \lambda\|z\|_{2}^{2} = \frac{1}{2} \substack{p \\ \sum \\ k=1}\lambda_{k}|x_{(k)}|^{2}$, where $A = I$ and $B = -I$. Therefore, ADMM is applicable to Equation (\ref{p3}). An~augmented Lagrangian form of Equation (\ref{p3}) can be defined as follows:
\vspace{-6pt}
\begin{align}\label{p4}
\begin{split}
\mathcal{L}_{p}(x,z,y) &= \frac{1}{2}\|Ax - b\|_{2}^{2}+ \frac{1}{2} \lambda \|z\|_{2}^{2} + y^{T}(x - z)\\
                        &+ \frac{\rho}{2} \|x - z\|_{2}^{2}
\end{split}
\end{align}
where $y \in \mathbb{R}^{p}$ is a~Lagrangian multiplier and $\rho>0$ denotes a~penalty parameter. Next, we apply ADMM to the augmented Lagrangian equation of Equation (\ref{p4}) (Reference~\cite{boyd2011distributed}, Section~3.1), which renders ADMM iterations as follows:
\begin{equation}\label{p5}
\begin{split}
x^{k+1} &:= \substack{argmin \\ x \in\ \mathbb{R}^{p}} \mathcal{L}_{p}(x,z^{k},y^{k}) \\
z^{k+1} &:= \substack{argmin \\ z \in\ \mathbb{R}^{p}} \mathcal{L}_{p}(x^{k+1},z,y^{k}) \\
y^{k+1} &:= y^{k} + \rho(x^{k+1} - z^{k+1})
\end{split}
\end{equation}

Proximal gradient methods are well known for solving convex optimization problems for which the objective function is the sum of a~smooth loss function and a~non-smooth penalty function \citep{schmidt2011convergence,parikh2014proximal,boyd2011distributed}. A well-studied example is $\ell_{1}$ regularized least squares \citep{bogdan2013statistical,tibshirani1996regression}. It should be noted that an~ordered $\ell_{1}$ norm is convex but not smooth. Therefore, these researchers used a~proximal gradient method. In contrast, we have employed an~ADMM method because ADMM can solve convex optimization problems for which the objective function is either the sum of a~smooth loss function and a~non-smooth penalty function or both loss and penalty function are smooth and ADMM also supports parallelism. In the ordered ridge regression, both loss and penalty function are smooth, whereas, in the ordered elastic net, loss~function is smooth and penalty function is non-smooth.
\subsection{Scaled Form}
\label{scaledform}
We can also define ADMM in scaled form by merging a~linear and a~quadratic term in augmented Lagrangian and then a~scaled dual variable, which is shorter and more appropriate. The scaled dual form of ADMM iterations in Equation (\ref{p5}) can be expressed as follows:
\begin{subequations}\label{p6}
\begin{align}
 x^{k+1} &:= \substack{argmin \\ x \in\ \mathbb{R}^{p}}(f(x) + (\rho/2)\|x - z^{k} + u^{k}\|_{2}^{2}) \label{p6:1} \\
 z^{k+1} &:= \substack{argmin \\ z \in\ \mathbb{R}^{p}}(g(z) + (\rho/2)\|x^{k+1} - z + u^{k}\|_{2}^{2}) \label{p6:2} \\
 u^{k+1} &:= u^{k} + x^{k} - z^{k} \label{p6:3}
\end{align}
\end{subequations}
where $u = \frac{1}{\rho}y$ and u is the scaled dual variable. Next, we can minimize the augmented Lagrangian in Equation (\ref{p4}) with respect to $x$ and $z$, successively. Minimizing Equation (\ref{p4}) with respect to 
$x$ becomes the $x$ subproblem of Equation (\ref{p6}a), and~it can be expressed as follows:
\begin{subequations}\label{p7}
\begin{align}
 \! \substack{min \\ x \in\ \mathbb{R}^{p}} \frac{1}{2}\|Ax - b\|_{2}^{2} + (\rho/2)\|x - z^{k} + u^{k}\|_{2}^{2}\nonumber\\
 = \substack{min \\ x \in\ \mathbb{R}^{p}} \frac{1}{2}\{x^{T}A^{T}Ax - 2b^{T}Ax\} + \frac{\rho}{2}\{x^{2} - 2(z^{k} - u^{k})^{T}x\} \label{p7:1}
 \! \intertext{\quad\quad After computing a~derivative of Equation (\ref{p7}a) with respect to $x$, then the setting of the derivative of $x$ becomes equal to zero. Notice that this is a~convex problem; therefore, it minimizes to solve the following linear system of Equation (\ref{p7}b):}
 \! \Leftrightarrow  A^{T}Ax + \rho x - b^{T}A - \rho (z^{k} - u^{k})^{T} = 0 \nonumber \\
 \! x^{k+1} = (A^{T}A + \rho \ast I)^{-1}(A^{T}b + \rho (z^{k} - u^{k}))^{T} \label{p7:2}
\end{align}
\end{subequations}

Minimizing problem Equation (\ref{p4}) w.r.t. $z$, we obtain Equation (\ref{p6}b), and~it results in the following $z$ subproblem:
\vspace{-6pt}
\begin{subequations}\label{p8}
\begin{align}
 \! \substack{min \\ z \in\ \mathbb{R}^{p}} \frac{1}{2}\lambda\|z\|_{2}^{2} + \frac{\rho}{2}\|x^{k+1} + u^{k} - z\|_{2}^{2}\nonumber\\
  = \substack{min \\ z \in\ \mathbb{R}^{p}} \frac{1}{2}\lambda_{k} z^{T}z + \frac{\rho}{2}\{(x^{k+1} + u^{k} - z)^{T}(x^{k+1} + u^{k} - z)\} \label{p8:1}
 \! \intertext{\quad\quad After computing a~derivative of Equation (\ref{p8}a) with respect to $z$, then the setting of the derivative of $z$ becomes equal to zero. Notice that this is a~convex problem; therefore, it minimizes to solve the following linear system of Equation (\ref{p8}b):}
 \! \Leftrightarrow  \frac{1}{2} 2 \lambda_{k} z + \frac{\rho}{2}\{2 z - 2(x^{k+1} + u^{k})^{T} \} = 0 \nonumber \\
 \! z^{k+1} = (\lambda_{k} + \rho \ast I)^{-1}\rho(x^{k+1} + u^{k})\label{p8:2}
\end{align}
\end{subequations}

Finally, the~multiplier (i.e., the~scaled dual variable $\emph{u}$) is updated in the following way:
\begin{align}
\label{p9}
u^{k+1} = u^{k} + (x^{k} - z^{k})
\end{align}

Optimality conditions: Primal and dual feasibility are essential and adequate optimality conditions for ADMM in Equation (\ref{p3})~\citep{boyd2011distributed}. Dual residual ($S^{k+1}$) and primal residual ($\gamma^{k+1}$) can be defined as follows:  
\begin{align}
\! &\text{Dual residual ($S^{k+1}$) at iteration $k+1$} &= \rho (z^{k} - z^{k+1}) \nonumber \\
\! &\text{Primal residual ($\gamma^{k+1}$) at iteration $k+1$} &= x^{k+1} - z^{k+1} \nonumber
\end{align}

Stopping criteria: The stopping criterion for an~ordered ridge regression is that primal and dual residuals must be small:
\begin{align}
\! &\|\gamma^{k}\|_{2} \leq \epsilon^{pri} \qquad \text{\tiny $where \quad \epsilon^{pri} = \sqrt{p}\epsilon^{abs} + \epsilon^{rel} max\{\|x^{k}\|_{2},\|z^{k}\|_{2}\}$} \nonumber \\
\! &\|S^{k}\|_{2} \leq \epsilon^{dual} \qquad \text{ \tiny $where \quad \epsilon^{dual} = \sqrt{n}\epsilon^{abs} + \epsilon^{rel} \|\rho \ast u^{k}\|_{2}$} \nonumber
\end{align}

We set $\epsilon^{abs} = 10^{-4}$ and $\epsilon^{rel} = 10^{-2}$. For further details about this choice, see Reference~\citep{boyd2011distributed}, Section~3).

\subsection{Over-Relaxed ADMM Algorithm}
\label{Over-relaxedform}
By comparing Equations (\ref{eq1:1}) and  (\ref{p3}), we can write Equation (\ref{p3}) in the over-relaxation form as follows:
\begin{align}
\! &\alpha x^{k+1} + (1 - \alpha)z^{k} \text{\tiny \quad /* where A = I, B = -I and c = 0 in our case */} \label{eq1:3}
\end{align}

Substituting $ x^{k+1}$ with Equation (\ref{eq1:3}) into $z$ of Equation (\ref{p8}b) and $u$ of Equation (\ref{p9}) updates the results in a~relaxation form. Algorithm \ref{alg:relaxedOrderedADMM} presents an~ADMM iteration for the ordered ridge regression of Equation (\ref{p3}).

  \vspace{6pt}

   \begin{algorithm}
    \caption{Over-relaxed ADMM for the ordered ridge regression}
    \label{alg:relaxedOrderedADMM}
    \begin{spacing}{1.6}
    \begin{algorithmic}[1]
        \STATE Initialize $x^{0}\in\ \mathbb{R}^{p}$, $z^{0}\in\ \mathbb{R}^{p}$, $u^{0}\in\ \mathbb{R}^{p}$, $\rho > 0$
        \WHILE{($\|\gamma^{k}\|_{2} \leq \epsilon^{pri} \quad \&\& \quad \|S^{k}\|_{2} \leq \epsilon^{dual}$)}
            \STATE $x^{k+1} \gets (A^{T}A + \rho \ast I)^{-1}(A^{T}b + \rho (z^{k} - u^{k}))^{T}$
            \STATE $\lambda_{k} \quad \gets \textit{SortedLambda}(\{\lambda_{k}\})$\Comment*[l]{\tiny refer to Algorithm \ref{alg:lambdaseq}}
            \STATE $z^{k+1} \gets (\lambda_{k} + \rho \ast I)^{-1}\rho(\alpha x^{k+1} + (1-\alpha)z^{k} + u^{k})$ \\
            \STATE $u^{k+1} \gets u^{k} + \alpha(x^{k+1} - z^{k+1}) + (1 - \alpha)(z^{k} - z^{k+1})$
        \ENDWHILE
    \end{algorithmic}
    \end{spacing}
 \end{algorithm}

 \vspace{12pt}

 {We observed that ADMM Algorithm \ref{alg:relaxedOrderedADMM} computes an~exact solution for each subproblem and that their convergence is guaranteed by existing ADMM theory~\citep{glowinski2008lectures,deng2016global,goldstein2014fast}. The most important and computationally intensive operation here is matrix inversion in line 3 of Algorithm \ref{alg:relaxedOrderedADMM}. Here, matrix A is high-dimensional ($p\gg n$) and $(A^{T}A + \rho \ast I)$ takes $\mathcal{O}(np^{2})$ and its inverse (i.e., $(A^{T}A + \rho \ast I)^{-1}$) takes $\mathcal{O}(p^{3})$. We compute $(A^{T}A + \rho \ast I)^{-1}$ and $A^{T}b$ outside loop; then, we are left with (inverse * $(A^{T}b + \rho (z^{k} - u^{k}))^{T}$), which is $\mathcal{O}(p^{2})$, while addition and subtraction take $\mathcal{O}(p)$. $(A^{T}A + \rho \ast I)^{-1}$ is also cacheable, so the complexity is just $\mathcal{O}(p^{3})$ + k * $\mathcal{O}(np^{2} + p)$ heuristically with k number of iteration.}\par

\textbf{Generating the ordered parameter $(\lambda_{k})$:}
As mentioned in the beginning, we set out to identify a~computationally tractable and adaptive solution. The regularizing sequences play a~vital role in achieving this goal. Therefore, we generated adaptive values of $(\lambda_{k})$ such that regressor coefficients are penalized according to their respective order. Our regularizing sequence procedure is motivated by the BHq procedure~\citep{benjamini1995controlling}. The BHq method generates $(\lambda_{k})$ sequences as follows:
\begin{align}
\! &\lambda_{BH^{(k)}} = \Phi^{-1}(1 - q * \frac{k}{2p}) \label{p13:1} \\
\! &\lambda_{k} = \lambda_{BH^{(k)}}\sqrt{1 + \frac{\substack{\sum \\ j<k}\lambda^{2}_{BH^{(j)}}}{n-k}} \label{p13:2}
\end{align}
where $k>0$, $\Phi^{-1}(\alpha)$ is $\alpha${th} quantile of a~standard normal distribution, and~$q$ is a~parameter, namely $q \in [0; 1]$. We started with $\lambda_{1} = \lambda_{BH^{(1)}}$ as an~initial value of the ordered parameter $(\lambda_{k})$.

Algorithm \ref{alg:lambdaseq} presents a~method for generating sorted $(\lambda_{k})$. The difference between lines 5 and 6 in Algorithm \ref{alg:lambdaseq} is that line 5 is for low-dimensional ($n \leq p$) data and that line 6 is for high-dimensional data ($p \gg n$). Finally, we used the ordered $(\lambda_{k})$ from Algorithm \ref{alg:lambdaseq} (i.e., the~adaptive value of $(\lambda_{k})$) in the ordered ridge regression of Equations (\ref{p3}) and (\ref{p4}) instead of ordinary $\lambda$. This makes the ordered $\ell_{2}$ adaptive and different from standard $\ell_{2}$.

\vspace{6pt}

\begin{algorithm}
   \caption{Sorted Lambda ($\{\lambda_{k}\})$}
   \label{alg:lambdaseq}
   \begin{spacing}{1.6}
   \begin{algorithmic}[1]
   \STATE Initialize $ q \in [0; 1], k>0, p, n \>\in N $
   \STATE  $\lambda_{1} \gets \lambda_{BH^{(1)}}$\Comment*[l]{\tiny $\lambda_{BH^{(1)}} \quad is \quad from \quad Equation (\ref{p13:1}) \quad where \quad k=1$}
   \FOR{$ k \in \{2,\dots,K\}$}
   \STATE  $\lambda_{BH^{(k)}} \gets \Phi^{-1}(1 - q * \frac{k}{2p})$
   \STATE  $\lambda_{k} \gets \lambda_{BH^{(k)}} \ast \sqrt{1+\frac{\substack{\sum \\ j<k}\lambda^{2}_{BH^{(j)}}}{p-k-1}}$\Comment*[l]{\tiny when n=p}
   \STATE  $\lambda_{k} \gets \lambda_{BH^{(k)}} \ast \sqrt{1+\frac{\substack{\sum \\ j<k}\lambda^{2}_{BH^{(j)}}}{2p-k-1}}$ \Comment*[l]{\tiny when n=2p}
   \ENDFOR
 \end{algorithmic}
 \end{spacing}
 \end{algorithm}
\subsection{The Ordered Elastic Net}
\label{elasticnet}
A standard $\ell_{2}$ (or an~ordered $\ell_{2}$) regularization is a~commonly used tool to estimate parameters for microarray datasets (strongly correlated grouping). However, a~key drawback of the $\ell_{2}$ regularization is that it cannot automatically select relevant variables because the $\ell_{2}$ regularization shrinks coefficient estimates closer but not exactly equal to zero (Reference~\cite{james2013introduction}, Chapter~6.2). 
On the other hand, a~standard $\ell_{1}$ (or an~ordered $\ell_{1}$) regularization can automatically determine relevant variables due to its sparsity property. However, the~$\ell_{1}$ regularization also has a~limitation. Especially when different variables are highly correlated, the~$\ell_{1}$ regularization tends to pick only a~few of them and to remove the remaining ones—even important ones that might be better predictors. To overcome the limitations of both $\ell_{1}$  and $\ell_{2}$ regularization, we proposed another method called an~ordered elastic net (the ordered $\ell_{1, 2}$ regularization or ADMM-O$\ell_{1, 2}$ or -O$\ell_{1, 2}$), similar to a~standard elastic net~\citep{zou2005regularization}, by combining the ordered $\ell_{2}$ regularization with the ordered $\ell_{1}$ regularization and the elastic net. By doing so,
the~ordered $\ell_{1, 2}$ regularization automatically selects relevant variables in a~way similar to the ordered $\ell_{1}$ regularization. In addition, it can select groups of strongly correlated variables. The key difference between the ordered elastic net and the standard elastic net is a~regularization term. We apply the ordered $\ell_{1}$ and $\ell_{2}$ regularization in the ordered elastic net instead of the standard $\ell_{1}$ and $\ell_{2}$ regularization. This~approach means that the ordered elastic net inherits the sparsity, grouping, and~adaptive properties of the ordered $\ell_{1}$  and $\ell_{2}$ regularization. We have also employed ADMM to solve the ordered $\ell_{1, 2}$ regularized loss minimization as follows:
\begin{align}
\label{enet1}
\Leftrightarrow \substack{min \\ x\ \in\ \mathbb{R}^{p}}\frac{1}{2}\|Ax - b\|_{2}^{2} + \alpha \lambda_{BH}\|x\|_{1} + \frac{1}{2} (1-\alpha)\lambda_{BH} \|x\|^{2}_{2} \nonumber
\intertext{\quad\quad For simplicity, let $\lambda_{1}=\alpha \lambda_{BH}$ and $\lambda_{2}=(1-\alpha) \lambda_{BH}$. The ordered elastic net becomes}
\Leftrightarrow \substack{min \\ x\ \in\ \mathbb{R}^{p}}\frac{1}{2}\|Ax - b\|_{2}^{2} + \lambda_{1}\|x\|_{1} + \frac{1}{2} \lambda_{2}\|x\|^{2}_{2} \nonumber
\intertext{\quad\quad Now, we can transform the above ordered elastic net equation into an~equivalent form of  Equation~(\ref{eq1:1}) by introducing an~auxiliary variable $z$.}
\Leftrightarrow \substack{min \\ x,z\ \in\ \mathbb{R}^{p}}\underbrace{\frac{1}{2}\|Ax - b\|_{2}^{2}}_{f(x)} + \underbrace{\lambda_{1}\|z\|_{1} + \frac{1}{2} \lambda_{2}\|z\|^{2}_{2}}_{g(z)}\ s.t.\ x-z = 0
\end{align}

We can minimize Equation (\ref{enet1}) w.r.t. $x$ and $z$ in the same way as we minimized the ordered $\ell_{2}$ regularization in Section \ref{ADMMorr} and Sections \ref{scaledform} and \ref{Over-relaxedform}. Therefore, we directly present the final results below without any details. The + sign means to select max (0,value).
\begin{equation}
\begingroup\makeatletter\def\f@size{10}\check@mathfonts
\def\maketag@@@#1{\hbox{\m@th\normalsize\normalfont#1}}%
\begin{array}{rl}
\label{enet2}
x^{k+1} &= (A^{T}A + \rho \ast I)^{-1}(A^{T}b + \rho (z^{k} - u^{k})) \\
z^{k+1} &= \Bigg(\frac{\rho(x^{k+1}+u^{k})-\lambda_{1}}{\lambda_{2}+\rho}\Bigg)_{+}-
\Bigg(\frac{-\rho(x^{k+1}+u^{k})-\lambda_{1}}{\lambda_{2}+\rho}\Bigg)_{+} \\
u^{k+1} &= u^{k} + (x^{k} - z^{k})
\end{array}
\endgroup
\end{equation}

\section{Experiments}
\label{experiment}
 A series of experiments were conducted on both simulated and real data to examine the performance of a~proposed method. In this section, first, a~concept to select a~correct sequence of $(\lambda_{k})$s is discussed. Second, an~experiment on synthetic data is presented that describes the convergence of the lasso, SortedL1, ADMM-O$\ell_{2}$, and~ADMM-O$\ell_{1, 2}$. Finally, the~proposed method is applied to a~real feature selection dataset. The performance of the ADMM-O$\ell_{1, 2}$ method is analyzed in comparison with state-of-the-art methods: the lasso and SortedL1. These two methods are chosen for comparison because they are very similar to the ADMM-O$\ell_{1, 2}$ method except they use the regular lasso and the ordered lasso, respectively, while ADMM-O$\ell_{1, 2}$ model employs the ordered $\ell_{1, 2}$ regularization with ADMM.

\textbf{Experimental setting:} The algorithms were implemented on Scala Spark{\texttrademark} with Scala code in both distributed and non-distributed versions. A distributed version of experiments was carried out in a~cluster of virtual machines with four nodes: one master and three slaves. Each node has 10 GB of memory, 8 cores, CentOS release 6.2, and~amd64: core-4.0-noarch. Apache spark{\texttrademark} 1.5.1 was deployed on it. The scholars also used IntelliJ IDEA 15 ULTIMATE as a~Scala editor, interactive build tool sbt version 0.13.8, and~Scala version 2.10.4. The standalone machine is a~Lenovo desktop running Windows 7 Ultimate with an~Intel{\texttrademark} Core{\texttrademark} i\textsubscript{3} Duo 3.20 GHz CPU and 4 GB of memory. The scholars used MATLAB{\texttrademark} version 8.2.0.701 running on a~single machine to draw all figures. The source codes of the lasso, SortedL1, and~ADMM-O$\ell_{1, 2}$ are available at References~\citep{lasso,SortedL1,ADMMOL2}, respectively.

\subsection{Adjusting the Regularizing Sequence $(\lambda_{k})$ for the Ordered Ridge Regression}\label{AnlaysisLambda}
\begin{figure}
\centering
\begin{minipage}[b]{0.49\textwidth}
  \centering
  \includegraphics[width=1.0\linewidth,height=1.0\linewidth]{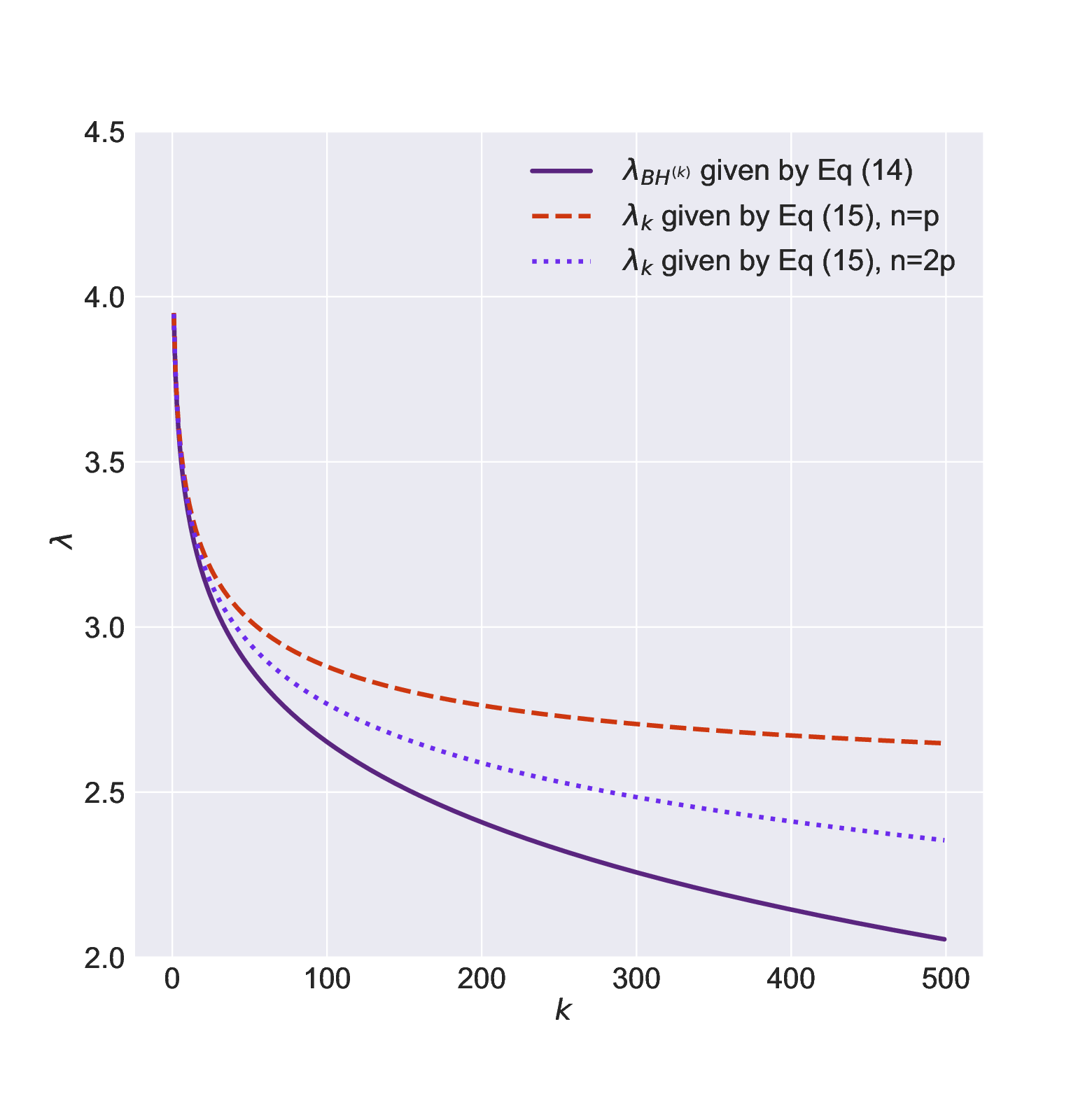}
  \\
  (\textbf{a}) $q = 0.4$
  \label{fig:lambdaA}
\end{minipage}
\begin{minipage}[b]{0.49\textwidth}
  \centering
  \includegraphics[width=1.0\linewidth,height=1.0\linewidth]{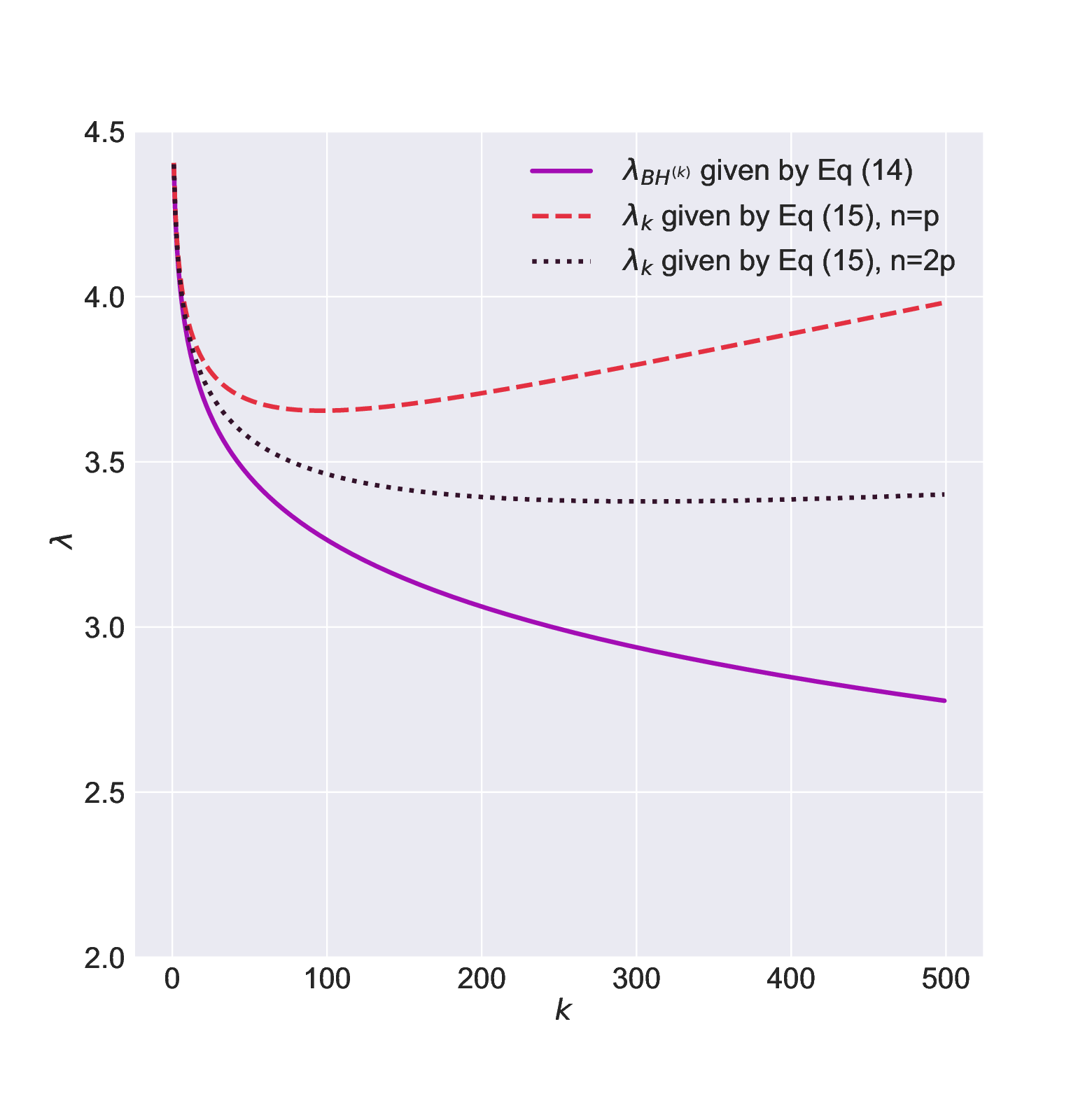}
  \\
  (\textbf{b}) $q = 0.055$
  \label{fig:lambdaB}
\end{minipage}
  \caption{Illustration of sequence $\{\lambda_{k}\}$ for p = 5000: The solid line of $\lambda_{BH^{(k)}}$ is given by Equation (\ref{p13:1}), while the dashed and dotted lines of $\lambda_{k}$ are given by Equation (\ref{p13:2}) for n = p and n = 2p, respectively.}
  \label{fig:lambda}
\end{figure}

Figure \ref{fig:lambda} was drawn using  Algorithm (\ref{alg:lambdaseq}), where p = 5000. As seen in Figure \ref{fig:lambda}, when the value of a~parameter ($q=0.4$) becomes larger, the~sequence $(\lambda_{k})$ decreases, while $(\lambda_{k})$ increases for a~small value of $q=0.055$. However, the~goal is to obtain a~non-increasing order of sequence $(\lambda_{k})$ by adjusting the value of $q$, which stimulates convergence. Here, adjusting means tuning the value of the parameter $q$ using the BHq procedure to yield a~suitable sequence $(\lambda_{k})$ such that it improves performance.

\vspace{-12pt}

\subsection{Experimental Results of Synthetic Data}
\begin{figure}
\centering
\begin{minipage}[t]{0.49\textwidth}
  \centering
  \includegraphics[width=1.0\linewidth,height=1.0\linewidth]{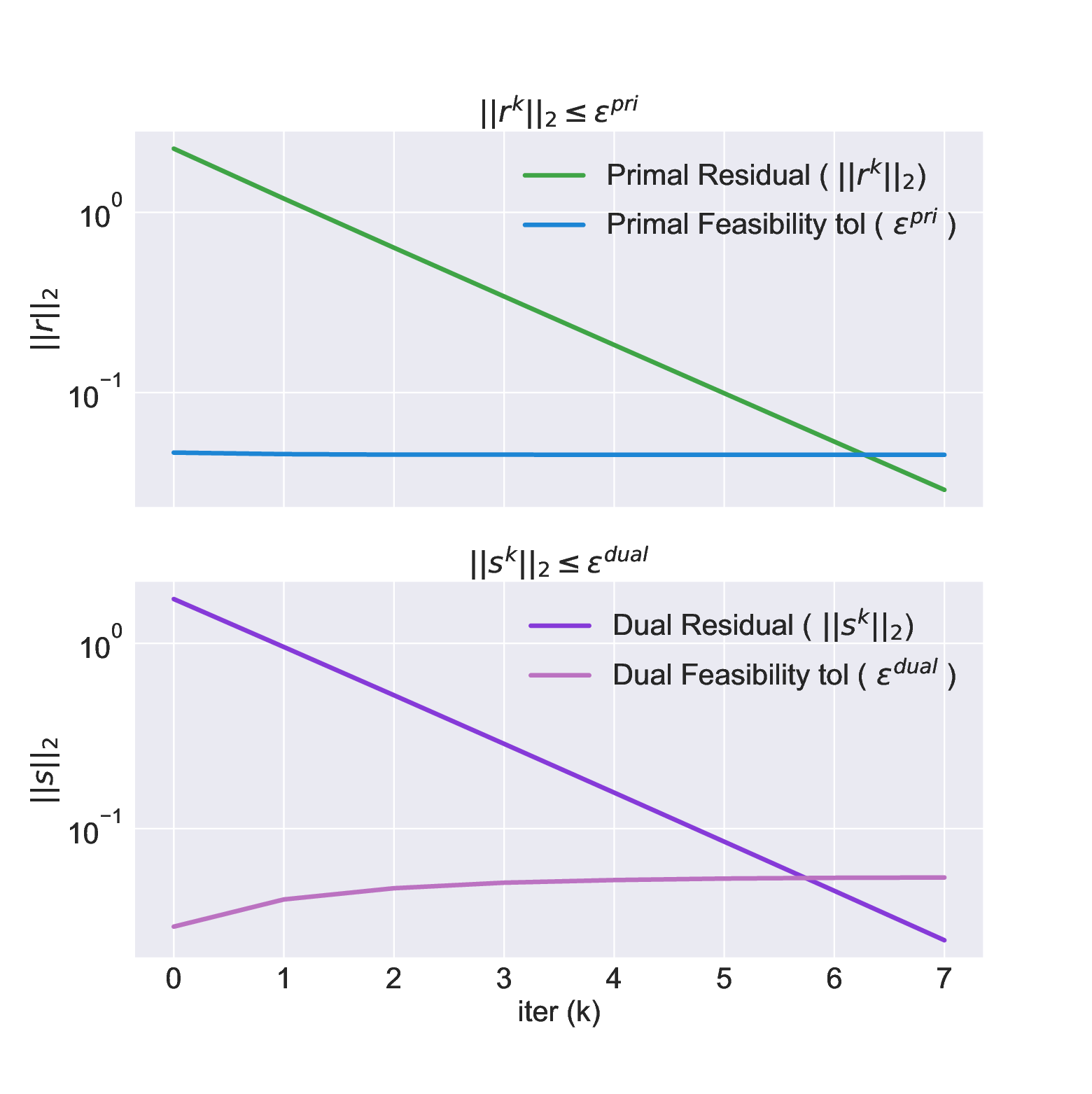}
  (\textbf{a}) The ordered $\ell_{2}$ regularization.
\end{minipage}
\begin{minipage}[t]{0.49\textwidth}
  \centering
  \includegraphics[width=1.0\linewidth,height=1.0\linewidth]{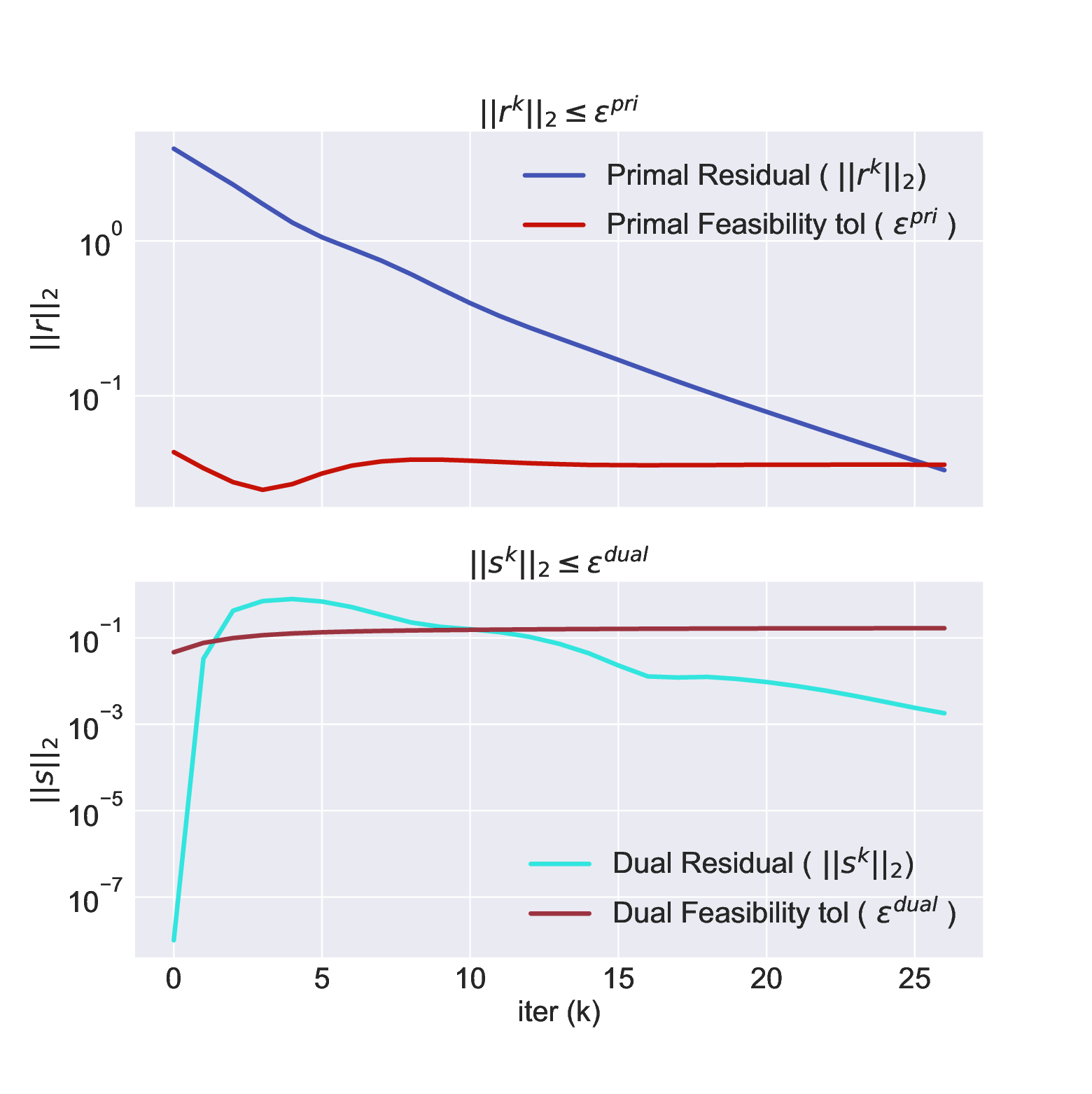}
  (\textbf{b}) The ordered $\ell_{1, 2}$ regularization.
\end{minipage}
\caption{Primal and dual residual versus primal and dual feasibility, respectively: Input synthetic data.}
\label{fig:OL1OL12}
\end{figure}

\begin{figure}
\centering
\begin{minipage}[t]{0.49\textwidth}
  \centering
    \includegraphics[width=1.0\linewidth,height=1.0\linewidth]{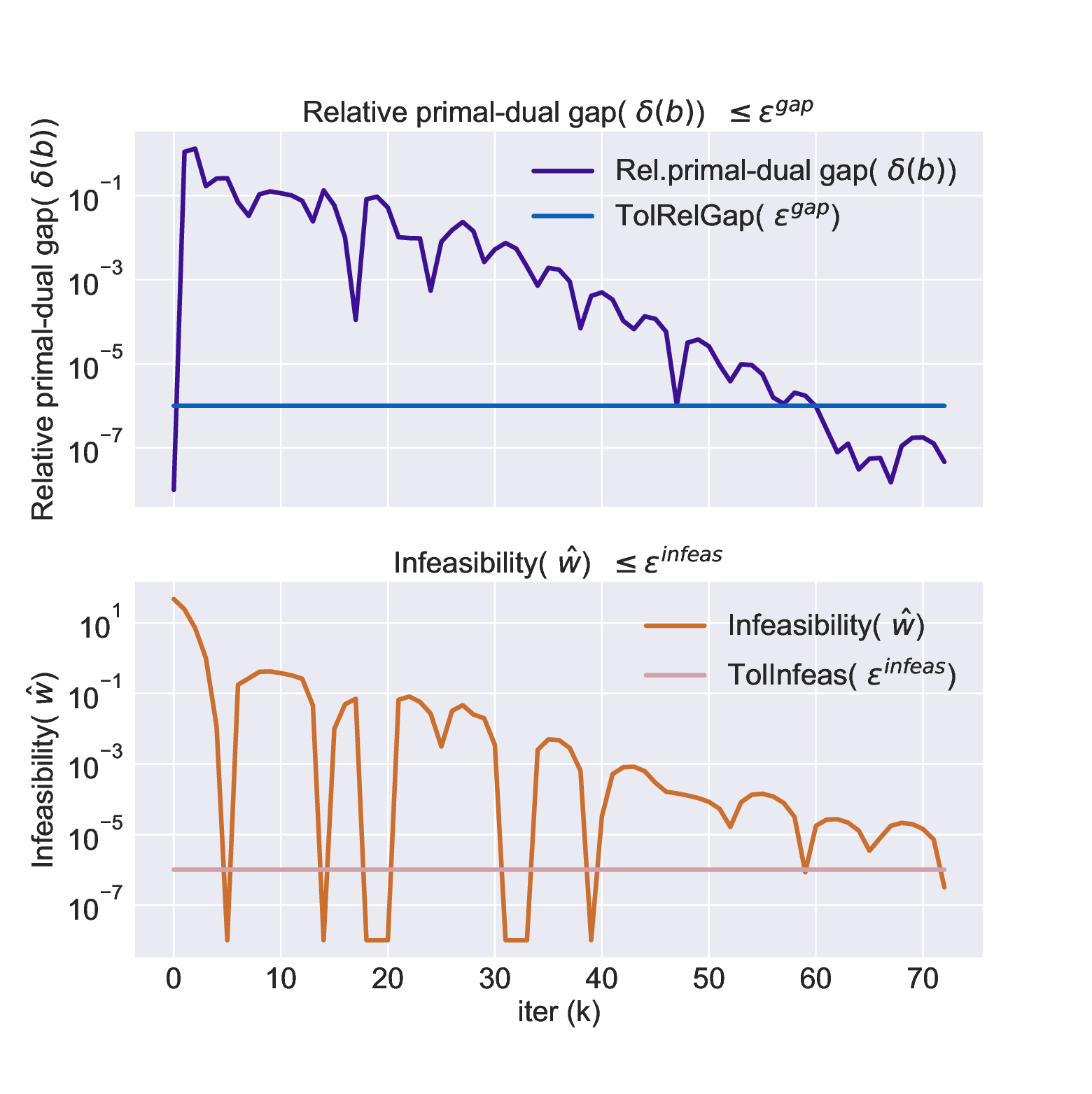}
  (\textbf{a}) The ordered $\ell_{1}$ regularization.
\end{minipage}
\begin{minipage}[t]{0.49\textwidth}
  \centering
  \includegraphics[width=1.0\linewidth,height=1.0\linewidth]{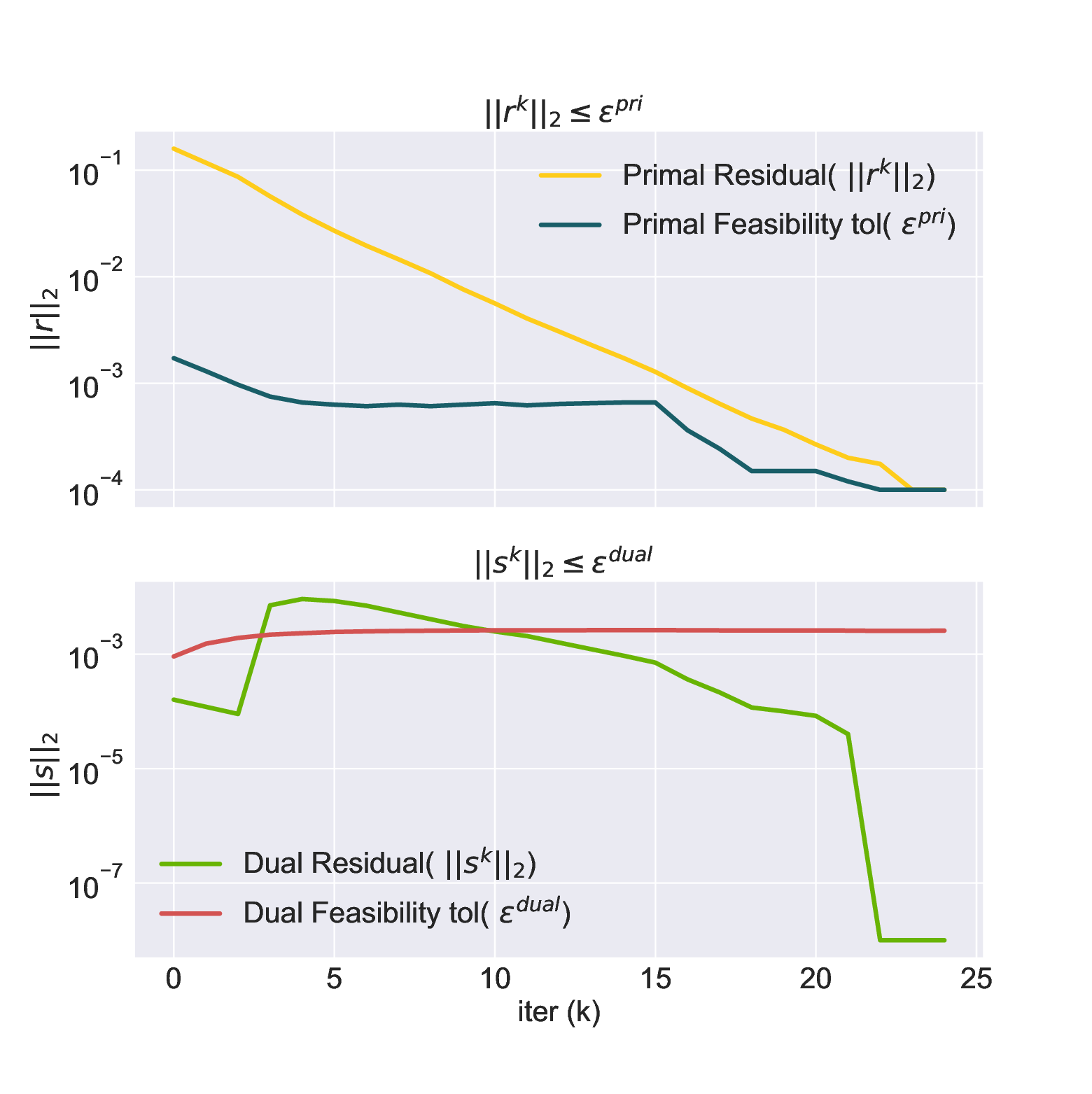}
  (\textbf{b}) The lasso.
\end{minipage}
\caption{(\textbf{a}) Relative primal--dual gap versus dual infeasibility, respectively, and~(\textbf{b}) primal and dual residual versus primal and dual feasibility, respectively: Input synthetic data.}
\label{fig:figOL1figL1}
\end{figure}

In this section, numerical examples show the convergences of ADMM-O$\ell_{1, 2}$, ADMM-O$\ell_{2}$, and~other methods. A tiny, dense example of an~ordered $\ell_{2}$ regularization is examined, where~the feature matrix A has n = 1500 examples and p = 5000 features. Synthetic data is generated as follows: create a~matrix $A$ and choose $A_{i,j}$ using $\mathcal{N}(0,1)$ and then normalize columns of the matrix $A$ to have the unit $\ell_{2}$ norm. $x^{0}\in \mathbb{R}^{p}$ is generated such that each sampled from $x^{0}\sim$ $\mathcal{N}(0,0.02)$ is a~Gaussian distribution. Label b is calculated as b = A*$x^{0}$ + v, where v $\sim$ $\mathcal{N}(0,10^{-3}*I)$, which is the Gaussian noise. A penalty parameter $\rho$ = 1.0, an~over-relaxed parameter $\alpha$ = 1.0, and~termination tolerances  $\epsilon^{abs} \leq 10^{-4}$ and  $\epsilon^{rel} \leq 10^{-2}$ are used. Variables $u^{0}\in \mathbb{R}^{p}$ and $z^{0}\in \mathbb{R}^{p}$ are initialized to be zero. $\lambda \in \mathbb{R}^{p}$ is a~non-increasing ordered sequence according to Section \ref{AnlaysisLambda} and Algorithm \ref{alg:lambdaseq}. Figure \ref{fig:OL1OL12}a,b indicates the convergence of ADMM-O$\ell_{2}$ and ADMM-O$\ell_{1, 2}$, respectively. Figure \ref{fig:figOL1figL1}a,b shows the convergence of the ordered $\ell_{1}$ regularization and the lasso, respectively. It can be seen from the Figures \ref{fig:OL1OL12} and \ref{fig:figOL1figL1} that the ordered $\ell_{2}$ regularization converges faster than all algorithms. The~ordered $\ell_{1}$, lasso, ordered $\ell_{1, 2}$, and~ordered $\ell_{2}$ take less than 80, 30, 30, and~10 iterations, respectively to converge. Dual is not guaranteed to be feasible. Therefore, a~level of infeasibility of dual is also needed to compute. A~numerical experiment terminates whenever both the infeasibility ($\hat{w}$) and relative primal-dual gap ($\delta(b)$) are less than equal to $\lambda(1) ($Tolinfeas ($\epsilon^{infeas}=10^{-6}$) and TolRelGap ($\epsilon^{gap}=10^{-6}$), respectively) the ordered $\ell_{1}$ regularization harness synthetic data provided by Reference~\citep{bogdan2013statistical}. The~same data is generated for the lasso as for the ordered $\ell_{2}$ regularization except for an~initial value of $\lambda$. For the lasso, set~$\lambda = 0.1*\lambda_{max}$, where~$\lambda_{max}=\|A^{T}*b\|_{\infty}$. The researchers also use 10-fold cross-validation (cv) with the lasso. For~further details about this step, see Reference~\citep{boyd2011distributed}.
\unskip
\subsection{Experimental Results of Real Data}
Variable selection difficulty arises when the number of features (p) are greater than the number of instances (n). The proposed method genuinely handles these types of issues. The practical application of the ADMM method is in many domains such as computer vision and graphics~\citep{liu2012tensor}, analysis of biological data~\citep{bien2013lasso,danaher2014joint}, and~smart electric grids~\citep{kraning2014dynamic,kekatos2012distributed}. A biological leukaemia dataset \citep{leukaemia_dataset} was used to demonstrate the performance of the proposed method. Leukemia is a~type of cancer that impairs the body's ability to build healthy blood cells. Leukemia begins in the bone marrow. There are many types of leukemia, such as acute lymphoblastic leukemia, acute myeloid leukemia, and~chronic lymphocytic leukemia. The following two types of leukemia are used in this experiment: acute lymphoblastic leukemia and acute myeloid leukemia. The leukemia dataset consists of 7129 genes and 72 samples~\citep{Golub1999MolecularCO}. Randomly split the data into training and test sets. In the training set, there are 38 samples, among which 27 are type I ALL (acute lymphoblastic leukemia) and 11 are type II AML (acute myeloid leukemia). The remaining 34 samples allowed us to test the prediction accuracy. The test set contains 20 type I ALL and 14 type II AML. The data were labeled according to the type of leukemia (ALL or AML). Therefore, before applying an~ordered elastic net, the~type of leukemia \mbox{(ALL = $-$1, or AML = 1)} is converted as a~($-$1, 1) response y. Predicted response $\hat{y}$ is set to 1 if $\hat{y}>0$; otherwise, it is set to $-$1. $\lambda \in \mathbb{R}^{p}$ is a~non-increasing, ordered sequence generated according to Section \ref{AnlaysisLambda} and Algorithm \ref{alg:lambdaseq}. For the regular lasso, $\lambda$ is a~single scalar value generated using Equation (\ref{p13:1}). $\alpha=0.1$ is used for the leukemia dataset. All other settings are the same as experiment with synthetic data.\par
\begin{figure}
\centerline{\includegraphics[scale=0.65]{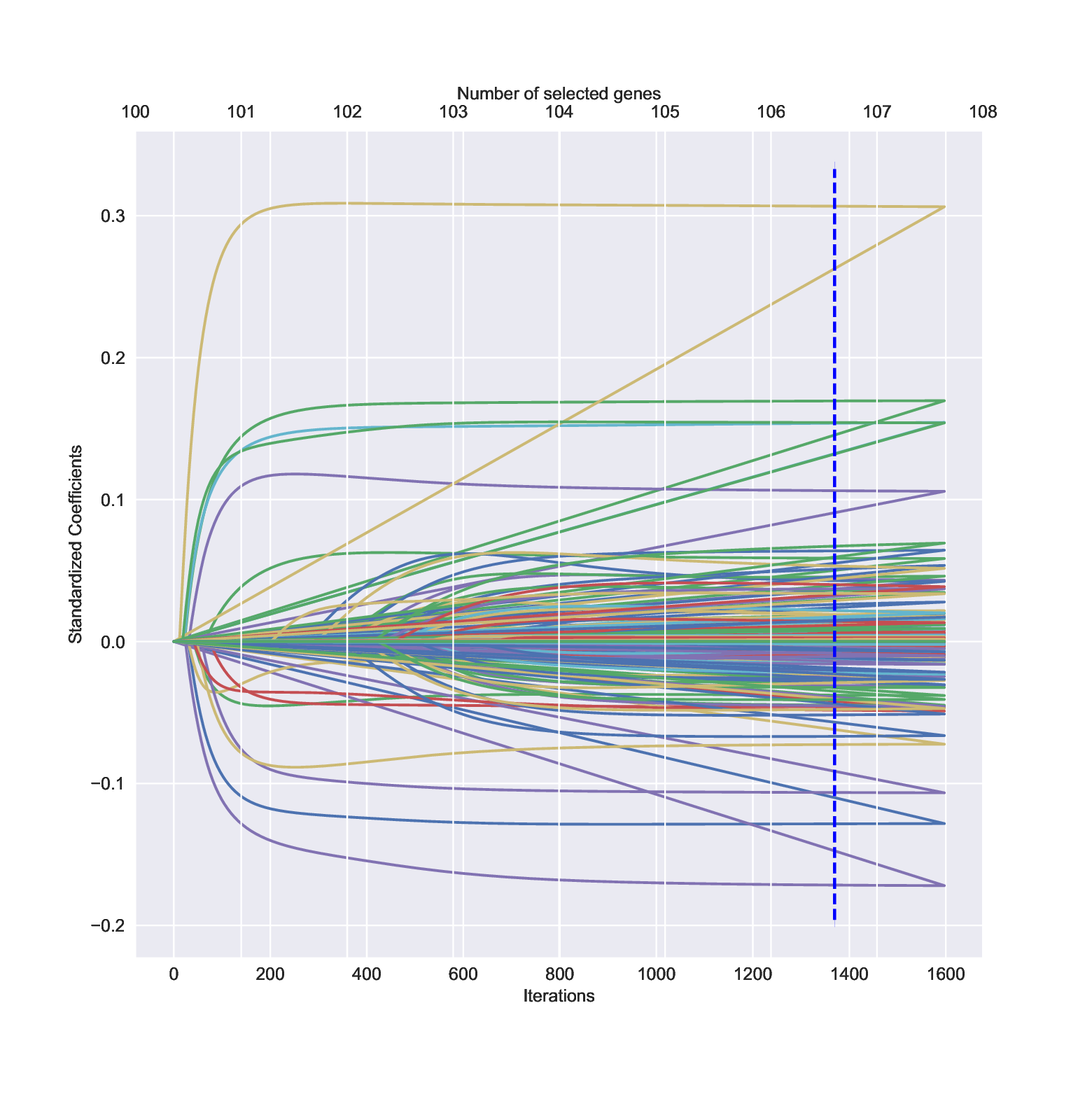}}
\caption{The ordered elastic net coefficients paths: selected genes (the number of nonzero coefficients) are shown on the top of the x-axis and corresponding iterations are shown on the bottom of the x-axis; the~optimal ordered elastic net model is given by the fit at an~average iteration of 1380.6 with an~average selected gene of 106.6 (indicated by a~dotted line). Input leukemia data.}
\label{fig:mse}
\end{figure}

Table \ref{tab:nzv} illustrates the experiment results of the leukemia dataset for different types of regularization. The lowest average mean square error (MSE) of regularization is the ordered $\ell_{2}$, followed by the ordered $\ell_{1, 2}$ and lasso, while the highest average MSE can be seen in the ordered $\ell_{1}$. Looking at Table \ref{tab:nzv} first, it is clear that the ordered $\ell_{2}$ converges the fastest among all the regularizations. The second fastest converging regularization is the ordered $\ell_{1, 2}$, while the slowest converging regularization is the ordered $\ell_{1}$. The ordered $\ell_{2}$ takes an~average iteration around 190 and an~average time around 0.15 s to converge. On the other hand, the~ordered $\ell_{1, 2}$,  the ordered $\ell_{1}$, and~lasso take average iterations around 1381, 10,000, and~10,000, respectively, and~average times around: 1.0, 14.0, and~5.0 s, respectively, to converge. It can also be seen from the data in Table \ref{tab:nzv} that the ordered $\ell_{2}$ selected all the variables but that the goal is to select only the relevant variables from strongly correlated, high-dimensional dataset. Therefore, the~ordered elastic net was proposed, which only selects relevant variables and discards irrelevant variables. As can be seen from  Table \ref{tab:nzv}, average MSE, time, and~iteration in the ordered $\ell_{1}$ regularization and lasso are significantly more than the ordered $\ell_{1,2}$ regularization, although an~average gene selection in the ordered $\ell_{1, 2}$ regularization is more than that of the ordered $\ell_{1}$ regularization and lasso. The ordered $\ell_{1}$ and lasso select averages around 84 and 7 variables, respectively, whereas the ordered $\ell_{1, 2}$ selects an~average around 107 variables. The~lasso performs poorly on the leukemia dataset. The reason for this is that strongly correlated variables are present in the leukemia dataset. In general, the~ordered elastic net performs better than the ordered $\ell_{1}$ and lasso. Figure \ref{fig:mse} shows the ordered elastic net solution paths and the variable selection results.
\begin{table}
  \centering
  \caption{Summary of variable selection in leukaemia dataset.}
  \label{tab:nzv}
\scalebox{.9}[.9]{\begin{tabular}{llllll}
\toprule
\textbf{q} & \textbf{Method} & \textbf{Test Error} & \textbf{\#Genes} & \textbf{Time} & \textbf{\#Iter} \\
\midrule
\multirow{4}{*}{0.1} & Lasso & 2.352941 & 6 & 5.459234 s & 10,000 \\
 & The ordered $\ell_{1}$ & 2.235294 & 56 & 16.208356 s & 10,000  \\
 & The ordered $\ell_{2}$ & 2.117647 & All & 0.176351 s & 216 \\
 & The ordered $\ell_{1,2}$ & 2.352941 & 109 & 1.391347 s & 2104 \\
\midrule
\multirow{4}{*}{0.2} & Lasso & 2.352941 & 6 & 5.419032 s & 10,000 \\
 & The ordered $\ell_{1}$ & 2.352941 & 65 & 14.990179 s & 10,000  \\
 & The ordered $\ell_{2}$ & 2.117647 & All & 0.167623 s & 197  \\
 & The ordered $\ell_{1,2}$ & 2.352941 & 107 & 1.046763 s & 1597 \\
\midrule
\multirow{4}{*}{0.3} & Lasso & 2.352941 & 6 & 5.394828 s & 10,000 \\
 & The ordered $\ell_{1}$ & 2.352941 & 85 & 15.477436 s & 10,000  \\
 & The ordered $\ell_{2}$ & 2.117647 & All & 0.140347 s & 185  \\
 & The ordered $\ell_{1,2}$ & 2.117647 & 108 & 0.820148 s & 1276  \\
\midrule
\multirow{4}{*}{0.4} & Lasso & 2.235294 & 7 & 5.470428 s & 10,000 \\
 & The ordered $\ell_{1}$ & 2.352941 & 90 & 12.206446 s & 10,000  \\
 & The ordered $\ell_{2}$ & 2.117647 & All & 0.135451 s & 178  \\
 & The ordered $\ell_{1,2}$ & 2.0 & 107 & 0.685255 s & 1055  \\
\midrule
\multirow{4}{*}{0.5} & Lasso & 2.0 & 8 & 5.387423 s & 10,000 \\
 & The ordered $\ell_{1}$ & 2.352941 & 126 & 13.226632 s & 10,000  \\
 & The ordered $\ell_{2}$ & 2.117647 & All & 0.126034 s & 172 \\
 & The ordered $\ell_{1,2}$ & 2.0 & 102 & 0.603964 s & 871 \\
\midrule
\multirow{4}{*}{Average} & Lasso & 2.2588234 & 6.6 & 5.426189 s & 10,000 \\
 & The ordered $\ell_{1}$ & 2.3294116 & 84.4 & 14.4218098 s & 10,000 \\
 & The ordered $\ell_{2}$ & 2.117647 & All & 0.1491612 s & 189.6 \\
 & The ordered $\ell_{1,2}$ & 2.1647058 & 106.6 & 0.9094954 s & 1380.6 \\
\bottomrule
\end{tabular}}
\end{table}
\unskip

\section{Conclusions}
\label{conclusions}
In this paper, we showed a~method for optimizing an~ordered $\ell_{2}$ problem under an~ADMM framework, called ADMM-O$\ell_{2}$. As an~implementation of ADMM-O$\ell_{2}$, the~ridge regression with the ordered $\ell_{2}$ regularization is shown. We also presented a~method for variable selection, called ADMM-O$\ell_{1, 2}$ which employs the ordered $\ell_{1}$ and $\ell_{2}$. We see the ordered $\ell_{1, 2}$ as a~generalization of the ordered $\ell_{1}$, which is shown as an~important tool for model fitting, feature selection, and~parameter estimation.\par

Experimental results show that the ADMM-O$\ell_{1, 2}$ method correctly estimates parameter, selects relevant variables, and~excludes irrelevant variables for microarray data. Our method is also computationally tractable, adaptive, and~distributed. The ordered $\ell_{2}$ regularization is convex and can be optimized efficiently with faster convergence rate. Additionally, we have shown that our algorithm has complexity $\mathcal{O}(p^{3})$ + k * $\mathcal{O}(np^{2} + p)$ heuristically, where k is the number of iterations. In future work, we plan to apply our method to other regularization models with complex penalties.


\acks{The work is funded by the National Key R\&D Program of China under Grant No. 2016QY02D0405, 973 Program of China under Grant No. 2014CB340401. Mahammad Humayoo is supported by CAS-TWAS fellowship. We gratefully acknowledge the useful comments of the anonymous referees. We would like to thank the editor for his continuous and quick support during the peer review process. We would also like to thank all students and teachers who supported us in this work. We discussed some issues with users in the online communities. Finally, we want to extend a big thank you to all the users of that online communities who participate in discussion and provide us valuable suggestion.}

\bibliographystyle{plain}
\bibliography{ADMM-OL2}

\appendix

\section{}\label{proofs}
\begin{proof}
\textbf{Positivity of Theorem \ref{theorem1}}
\begin{align}
J_{\lambda}(x) &= \lambda\|x\|^{2}_{2} = \substack{p \\ \sum \\ k=1}\lambda_{k} x_{(k)}^{2}= \substack{p \\ \sum \\ k=1} (\sqrt{\lambda_{k}}x_{(k)})^{2} \nonumber\\
\shortintertext{\quad\quad Take the square root on both side for the following:}
\|\sqrt{\lambda}x\|_{2}&=  \sqrt{\substack{p \\ \sum \\ k=1}(\sqrt{\lambda_{k}}x_{(k)})^{2}} \nonumber\\
\shortintertext{\quad\quad For $\forall x\in\mathbb{R}^{p}$, hold Equation (\ref{lambdaseq}); thus, $(\lambda_{1..p})$ is positive and $(\lambda_{1..p})\neq0$. $\|x\|_{2}$ will never be negative because of the square of $x$.}
\|\sqrt{\lambda}x\|_{2}&= \sqrt{\substack{p \\ \sum \\ k=1}(\sqrt{\lambda_{k}}x_{(k)})^{2}} \geq 0 \nonumber
\shortintertext{\quad\quad If $x=0$, then we have the following:}
\|\sqrt{\lambda}x\|_{2}&= \sqrt{\substack{p \\ \sum \\ k=1}(\sqrt{\lambda_{k}}.0)^{2}}=0
\shortintertext{\quad\quad If $\|x\|_{2}=0$, then we have the following:}
\|\sqrt{\lambda}x\|_{2}&= \sqrt{\substack{p \\ \sum \\ k=1}(\sqrt{\lambda_{k}}x_{(k)})^{2}}=0\Rightarrow x_{(k)}=0 \forall k \Rightarrow x=0 \nonumber\end{align}

{Since $(\lambda_{k})\neq0$, $\|x\|_{2}$ will be only zero if and only if $x=0$.}
\end{proof}
\begin{proof}
\textbf{Homogeneity of Theorem \ref{theorem1}}
\begin{align}
\shortintertext{\quad\quad The first two steps are the same as that of the proof of positivity of Theorem \ref{theorem1}; then, we have the following:}
\|\sqrt{\lambda}x\|_{2}&= \sqrt{\substack{p \\ \sum \\ k=1}(\sqrt{\lambda_{k}})^{2}x_{(k)}^{2}}= |\sqrt{\lambda}|\sqrt{\substack{p \\ \sum \\ k=1}x_{(k)}^{2}} \nonumber\\
\|\sqrt{\lambda}x\|_{2}&= |\sqrt{\lambda}|\|x\|_{2} \nonumber\\
\shortintertext{where ($c=\sqrt{\lambda}$). Then, we have the following:}
\|cx\|_{2}&= |c|\|x\|_{2} \nonumber
\end{align}
\end{proof}

\begin{proof}
\textbf{Triangle inequality of Theorem \ref{theorem1}}
\begin{align}
\shortintertext{\quad\quad The first two steps are the same as that of the proof of positivity of Theorem \ref{theorem1}. Now, with $x+y$ in place of $x$, we have the following:}
\|\sqrt{\lambda}(x+y)\|_{2}&= \sqrt{\substack{p \\ \sum \\ k=1}(\sqrt{\lambda_{k}}x_{(k)}+\sqrt{\lambda_{k}}y_{(k)})^{2}} \nonumber\\
\|\sqrt{\lambda}(x+y)\|_{2}&= \sqrt{\substack{p \\ \sum \\ k=1}(\sqrt{\lambda_{k}}x_{(k)})^{2}+\substack{p \\ \sum \\ k=1}(\sqrt{\lambda_{k}}y_{(k)})^{2}+2\substack{p \\ \sum \\ k=1}\lambda_{k}x_{(k)}y_{(k)}} \nonumber
\shortintertext{\quad\quad From Cauchy--Schwarz inequality, $x.y\leq\|x\|.\|y\|$, we have the following:}
\|\sqrt{\lambda}x+\sqrt{\lambda}y\|_{2}&\leq \sqrt{\|\sqrt{\lambda}x\|^{2}+\|\sqrt{\lambda}y\|^{2}+2\|\sqrt{\lambda}x\|.\|\sqrt{\lambda}y\|} \nonumber\\
\|\sqrt{\lambda}x+\sqrt{\lambda}y\|_{2}&\leq \sqrt{(\|\sqrt{\lambda}x\|+\|\sqrt{\lambda}y\|)^{2}}\nonumber\\
\|\sqrt{\lambda}x+\sqrt{\lambda}y\|_{2}&\leq (\|\sqrt{\lambda}x\|+\|\sqrt{\lambda}y\|)\nonumber
\shortintertext{where $x=\sqrt{\lambda}x$ and $y=\sqrt{\lambda}y$. Then, we have the following:}
\|x+y\| &\leq \|x\|+\|y\|\nonumber
\end{align}
\end{proof}

\begin{proof}
\textbf{Corollary \ref{coroll1}}
\begin{align}
\shortintertext{\quad\quad From Equation (\ref{lfunc}):}
J_{\lambda}(x) &= \lambda_{1} x_{(1)}^{2} + \lambda_{2} x_{(2)}^{2} + \dots=+ \lambda_{p} x_{(p)}^{2}\nonumber
\shortintertext{\quad\quad All $\lambda_{k}$ take on an~equal positive value, i.e., $\lambda_{1}=\lambda_{2}=\dots=\lambda_{p}$}
J_{\lambda}(x) &= \lambda x_{(1)}^{2} + \lambda x_{(2)}^{2} +\dots=+ \lambda x_{(p)}^{2}\nonumber\\
J_{\lambda}(x) &= \lambda (x_{(1)}^{2} + x_{(2)}^{2} +\dots+ x_{(p)}^{2})\nonumber\\
J_{\lambda}(x) &= \lambda \substack{p \\ \sum \\ k=1}x_{(k)}^{2}\nonumber\\
J_{\lambda}(x) &= \lambda \|x\|^{2}_{2}\nonumber\end{align}
{where $\lambda$ is positive scalar.}
\end{proof}
\begin{proof}
\textbf{Corollary \ref{coroll2}}
\begin{align}
\shortintertext{\quad\quad The first two steps are the same as that of the proof of positivity of Theorem \ref{theorem1}. When $\lambda_{2}$ = \dots = $\lambda_{p}$ = 0, we get the first term only and $p=1$. The remaining terms are zero when $p>1$.}
\|\sqrt{\lambda}x\|_{2}&=  \sqrt{\substack{p=1 \\ \sum \\ k=1}(\sqrt{\lambda_{k}}x_{(k)})^{2}} \nonumber\\
\|\sqrt{\lambda}x\|_{2}&=  \sqrt{(\sqrt{\lambda_{1}}x_{(1)})^{2}} \nonumber\\
\|\sqrt{\lambda}x\|_{2}&=  |\sqrt{\lambda_{1}}x_{(1)}| \nonumber\\
\shortintertext{where $x_{1}=\sqrt{\lambda_{1}}x_{(1)}$ and $\|x\|_{\infty}=max|x_{k}|$. Then, we have the following:}
\|x\|_{2}&= \|x\|_{\infty} \nonumber
\end{align}
\end{proof}

\end{document}